\documentclass{article}

\usepackage[preprint]{neurips_2022}

\usepackage[utf8]{inputenc} 
\usepackage[T1]{fontenc}    

\usepackage[dvipsnames,table]{xcolor}
\usepackage[pagebackref=false,breaklinks=true,letterpaper=true,colorlinks,urlcolor = black,  citecolor = OliveGreen, bookmarks=false]{hyperref}

\usepackage{url}            
\usepackage{booktabs}       
\usepackage{amsfonts}       
\usepackage{nicefrac}       
\usepackage{microtype}      
\usepackage{xcolor}         

\usepackage{makecell}

\usepackage{dsfont}

\usepackage{comment}
\usepackage{amsmath}

\usepackage{multirow}

\usepackage{threeparttable}

\usepackage{enumitem}

\usepackage{bbding}

\usepackage{natbib}
\setcitestyle{numbers,square}

\usepackage[pdftex]{graphicx}

\usepackage{wrapfig}
\usepackage{gensymb}
\usepackage{xparse} 
\usepackage{caption}
\usepackage{multicol}
\usepackage{subcaption}
\usepackage{caption}
\usepackage[ruled,vlined]{algorithm2e}
\SetKw{Break}{break}

\newcommand{\changed}[1]{}

\title{Pro-tuning: Unified Prompt Tuning for Vision Tasks}

\author{
	Xing Nie$^{1,2}$,
	Bolin Ni$^{1,2}$,
	Jianlong Chang$^3$,
	Gaomeng Meng$^{1,2,4}$,
	Chunlei Huo$^1$\thanks{Corresponding author.}, \\
	Zhaoxiang Zhang$^{1,4}$,
	Shiming Xiang$^1$,
	Qi Tian$^3$,
	Chunhong Pan$^1$
	\\[0.2cm]
	\small$ ^1$ National Laboratory of Pattern Recognition, Institute of Automation, Chinese Academy of Sciences. \\
	\small$ ^2$ School of Artificial Intelligence, University of Chinese Academy of Sciences.
	\small$ ^3$ Huawei Cloud \& AI. \\
	\small$ ^4$ CAS Centre for Artificial Intelligence and Robotics, HK Institute of Science and Innovation.\\
	\small\texttt{Email:\;niexing2019@ia.ac.cn,\;\{gfmeng,\;clhuo\}@nlpr.ia.ac.cn} \\
}

\begin{document}

	\maketitle

	\begin{abstract}
In computer vision, fine-tuning is the de-facto approach to leverage pre-trained vision models to perform downstream tasks. However, deploying it in practice is quite challenging, due to adopting parameter inefficient global update and heavily relying on high-quality downstream data. Recently, prompt-based learning, which adds a task-relevant prompt to adapt the downstream tasks to pre-trained models, has drastically boosted the performance of many natural language downstream tasks. In this work, we extend this notable transfer ability benefited from prompt into vision models as an alternative to fine-tuning. To this end, we propose parameter-efficient \textbf{Pro}mpt \textbf{tuning}~(Pro-tuning) to adapt frozen vision models to various downstream vision tasks. The key to Pro-tuning is prompt-based tuning, \textit{i.e.}, learning task-specific vision prompts for downstream input images with the pre-trained model frozen. By only training a few additional parameters, it can work on diverse CNN-based and Transformer-based architectures. Extensive experiments evidence that Pro-tuning outperforms fine-tuning in a broad range of vision tasks and scenarios, including image classification~(generic objects, class imbalance, image corruption, adversarial robustness, and out-of-distribution generalization), and dense prediction tasks such as object detection and semantic segmentation. 
	\end{abstract}

	\section{Introduction}
	
\begin{wrapfigure}{r}{7cm}
		\vspace{0mm}
		\centering
		\includegraphics[width=0.5\textwidth]{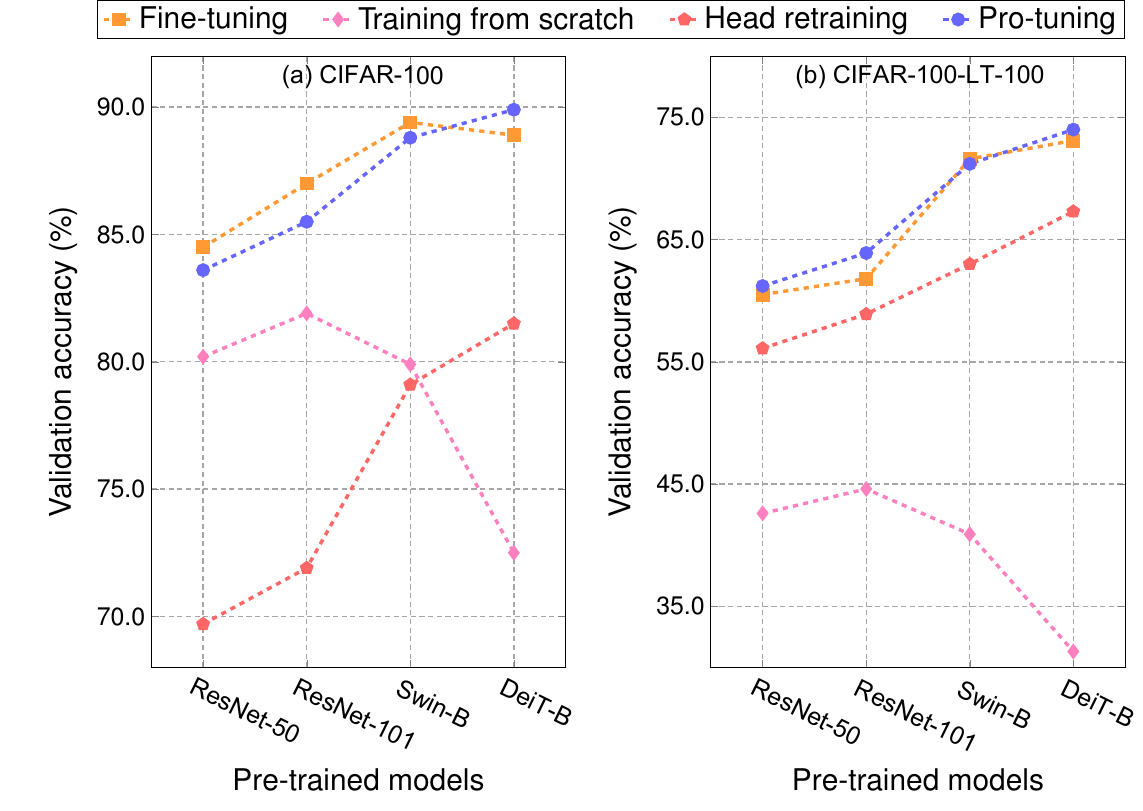}
		\caption{Comparison between the proposed Pro-tuning and other tuning paradigms under various pre-trained vision models. CIFAR-100-LT-100: Long-tailed CIAFR-100 with imbalanced ratio 100.}
		\vspace{-5mm}\label{fig:teaser1}
	\end{wrapfigure}

	Fine-tuning is the dominant technique for using pre-trained vision models to perform downstream tasks, which has contributed towards significant advances in a wide variety of computer vision problems~\cite{anderson2018bottom, lin2017feature, liu2021swin, long2015fully, touvron2021training, vinyals2015show}. Its standard practice is first to pre-train a general-purpose model on a large-scale dataset (\textit{e.g.}, ImageNet~\cite{deng2009imagenet}), then fine-tune the entire pre-trained model on downstream tasks. Although the achievements in the literature are brilliant, the fine-tuned model is still quite intractable to handle practical downstream vision applications for two main reasons. First, it requires updating and storing all model parameters separately for each downstream task, which is prohibitively expensive for the current ever-increasing vision model capacity. Second, its effectiveness heavily relies on high-quality downstream data, however, realistic data encountered in practical scenarios probably contains various negative distractors. In Figure~\ref{fig:teaser1}, we provide an empirical study of fine-tuning along with other popular tuning strategies. Though achieving decent accuracy on CIFAR-100~\cite{krizhevsky2009learning}, fine-tuning encounters devastating damages on various pre-trained vision models under distribution perturbations, \textit{e.g.}, more than 12\% validation accuracy drop on long-tailed CIFAR-100~\cite{cao2019learning} with imbalanced ratio 100 over their performances on the original CIFAR-100. 
	
	Recently, prompt-based learning has made waves in the natural language processing~(NLP) community by demonstrating astounding transfer performance on myriad downstream tasks~\cite{brown2020language, gao2020making, petroni2019language, radford2018improving, radford2019language, schick2020exploiting, talmor2020olmpics}. Typically, prompt is a task-relevant description prepended to the downstream input to induce the downstream task to the pre-trained model. Its key idea is to reformulate the downstream task aided by an appropriate prompt design, making it close to those solved during the original pre-training, rather than fine-tuning pre-trained models to adapt to downstream tasks. 
	Following this ideology, vision-language pre-training~\cite{jia2021scaling, kamath2021mdetr, radford2021learning, zhou2021learning} has been gradually developed to put visual categories into a text input as prompt, in order to diametrically generate visual concepts from natural language. Though achieving remarkable performance on many downstream vision tasks even without fine-tuning the language model, their prompt tuning strategies are tailored for the multimodal model, inapplicable to the pre-trained vision model. A question naturally arises:~can we devise a lightweight prompt tuning paradigm specifically for the pre-trained vision models to boost downstream vision tasks?

	In this spirit, we introduce \textit{Pro-tuning}, a unified tuning paradigm for a variety of vision models, including convolutional neural networks (CNNs) and vision transformer models. The key to Pro-tuning is prompt-based tuning, \textit{i.e.}, only learning task-specific vision prompts for downstream input images while freezing the pre-trained vision model. Specifically, Pro-tuning constructs lightweight prompt blocks to generate task-specific discriminative prompts for each downstream input image. By blending intermediate feature maps of each input with the learned prompts, Pro-tuning can generate a compact and robust downstream model to adapt tuning demands on various vision tasks, while only training lightweight additional parameters per downstream task.

	We demonstrate that Pro-tuning is a generic tuning framework for CNN-based and Transformer-based vision models, which can surpass fine-tuning on various recognition tasks of object detection, semantic segmentation, and image classification under in a variety of practical scenarios, including generic objects, class imbalance, image corruption, adversarial robustness, and out-of-distribution generalization. Figure~\ref{fig:teaser1} illustrates the quantitative results of Pro-tuning and other tuning paradigms on two image classification datasets. In particular, on the original CIFAR-100~\cite{hendrycks2021many}, Pro-tuning achieves a superior result compared with fine-tuning under DeiT-B~\cite{touvron2021training}, which brings in a significant gain of +1.0\% validation accuracy with only 27.1x fewer trainable parameters. On the more challenging long-tailed CIFAR-100~\cite{cao2019learning} with imbalanced ratio 100, Pro-tuning brings +0.6\% and +2.1\% validation accuracy gains over fine-tuning under Swin-S~\cite{liu2021swin} and ResNet-101~\cite{he2016deep}, which reduces the trainable parameters by up to 31x and 11x, respectively. Regarding object detection, the proposed Pro-tuning obtains comparable performance to full fine-tuning~(42.3 AP \textit{v.s.}, 42.2 AP) under HTC~\cite{chen2019hybrid} on COCO val2017 while reducing the trainable parameters by 25.5\%~(80.0M \textit{v.s.}, 60.4M). On ADE20K semantic segmentation, it brings +0.6 mIoU gains~(41.3 mIoU \textit{v.s.}, 40.7 mIoU) over full fine-tuning under UperNet~\cite{xiao2018unified} with the significant reduction of trainable parameters by 29.9\%~(46.6M \textit{v.s.}, 66.5M).

	The key contributions can be summarized as follow:
	\begin{itemize}
		\setlength{\itemsep}{0ex}
		\item A parameter-efficient vision tuning paradigm, \textit{i.e.}, Pro-tuning, is proposed. This prompt-based tuning specifically for vision models can adapt frozen pre-trained models to downstream tasks with broad distributions. 
		\item The proposed Pro-tuning can cooperate with various vision models including CNNs and vision transformers. As a plug-and-play technique, it is a capable general-purpose tuning framework with minor additional computational costs.
		\item Extensive experiments demonstrate that Pro-tuning exceeds fine-tuning on sixteen challenging benchmarks across three vision tasks, including image classification under five scenarios, and dense prediction tasks such as object detection and semantic segmentation. 
	\end{itemize}

	\section{Related Work}
	\subsection{Model Fine-Tuning}
	In computer vision, a series of techniques have been proposed to adapt general-purpose vision models to downstream tasks~\cite{chen2019catastrophic, li2019delta, xuhong2018explicit, yosinski2014transferable, you2020co, you2021logme}. However, the dominant techniques are still fine-tuning the entire moderately-sized pre-trained vision model~(\textit{e.g.}, ResNet~\cite{he2016deep} and DeiT-B~\cite{touvron2021training}) on the downstream data. Recently, fine-tuning techniques have enabled remarkable progress in the NLP community~\cite{houlsby2019parameter, lin2020exploring, pfeiffer2020adapterfusion, rebuffi2017learning}. Notably, the flagship systems like GPT-3~\cite{brown2020language}~(composed of 175B parameters) achieve remarkable performance across a wide range of downstream tasks even without requiring additional task-specific parameters. Recent work~\cite{kornblith2019better, radford2021learning, tsimpoukelli2021multimodal, you2020co} empirically validates that larger pre-trained models have a tendency to achieve better transfer performance, thus an important ingredient in leading to such difference between computer vision and NLP could be the evident increasing scale of the pre-trained language model. In this work, we develop a parameter-efficient tuning paradigm for substantially exploiting the knowledge in frozen pre-trained vision models. Our focus is on exploring how far we can push without the advantages of large-scale pre-trained models.

	\subsection{Prompt-based Learning}
	Prompt-based learning~\cite{liu2021pre} is proposed and fueled by the GPT series~\cite{brown2020language, radford2018improving, radford2019language} in the field of NLP. This is an approach of adding task-relevant descriptions to the downstream input to help the pre-trained model understand the downstream tasks, rather than only adapting the pre-trained model to downstream tasks by fine-tuning. GPT-3 pioneers this route by treating each downstream task as a masked language modeling problem, where the model directly generates a textual response in prompt. Subsequently, numerous researches are devoted to devising efficient prompt strategies for mining knowledge from pre-trained language models~\cite{lester2021power, gao2020making, li2021prefix, schick2020exploiting, talmor2020olmpics}. Motivated by GPT-3, vision-language models such as CLIP~\cite{radford2021learning} and ALIGN~\cite{jia2021scaling} diametrically generate vision concepts from natural language by training a large contrastive learning model over a large number of image-text pairs, where the key is to place visual categories into a text input as prompt. They achieve impressive performance on a wide range of vision tasks without any fine-tuning. However, prompt-based tuning methods in vision-language models are tailored for the multimodal model, and inapplicable to the pre-trained vision model. To fill this gap, our work aims to develop a parameter-efficient prompt tuning paradigm specifically for vision models, in order to adapt the frozen pre-trained vision model to downstream tasks on a broad distribution.

	\section{Pro-tuning}
	In this section, we describe the integration of Pro-tuning into pre-trained vision models with diverse network architectures, including CNNs and vision transformer models.
	\subsection{Overall Architecture}
	\label{section3.1}
\begin{wrapfigure}{r}{7cm}
		\vspace{-3mm}
		\centering
		\includegraphics[width=0.5\textwidth]{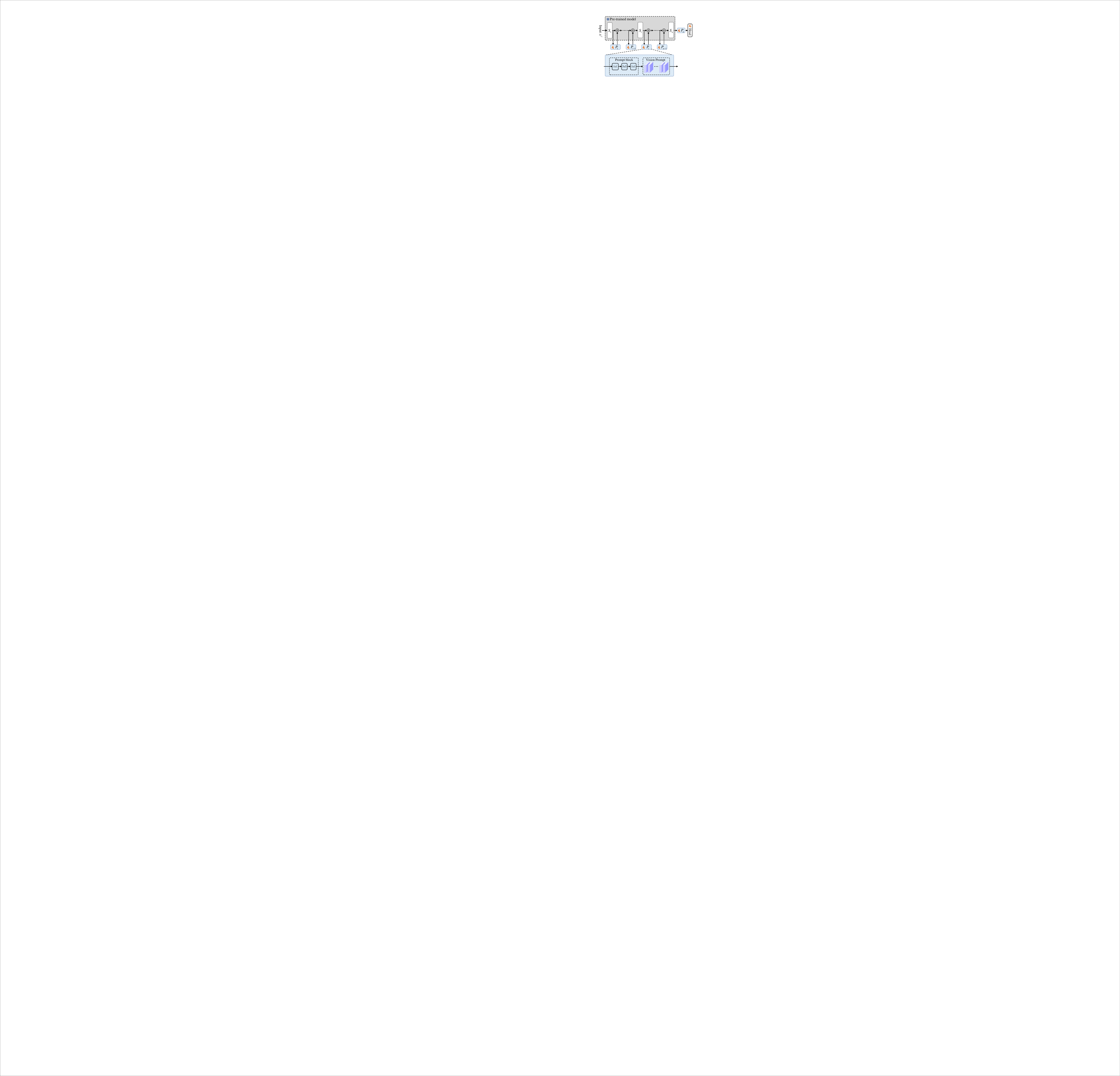}
		\caption{An overview of our proposed Pro-tuning.}
		\vspace{-4mm}\label{fig:overview}
	\end{wrapfigure}

	Pro-tuning is a method for adapting a pre-trained vision model to downstream tasks without changing its weight. To this end, we introduce lightweight prompt blocks to learn task-specific prompts for each downstream input, while freezing the pre-trained model. Inspired by studies~\cite{islam2021broad, lin2017refinenet, yosinski2014transferable}, feature representations of different levels contribute to the generalization performance of the network, especially for low-level and mid-level representations. To sufficiently exploit semantic information from diverse levels to enrich the feature space, we build multiple stage-wise prompt blocks, which can be simply plugged into a given frozen pre-trained vision model for quickly adapting to a new task. Specifically, this process can be computed through four simple steps: 1) feeding each downstream input image into the frozen pre-trained model; 2) adopting intermediate feature maps from different stages to train stage-wise prompt blocks for producing prompts; 3) blending the learned prompts with the corresponding feature maps as input to the next stage of the frozen pre-trained model; 4) appending a fully-connected layer to top layers of the pre-trained model for the final prediction. In this way, the task-specific vision prompts can be obtained, maximally mitigating the gap between pre-training and tuning on downstream tasks. It is worth noting that vision models typically divide the whole network architectures into multiple sub-layers as stages, for either transformer-based or CNN-based models. To ensure the generality of our method, we follow their original concepts of stage partitioning. Intuitively, our Pro-tuning is illustrated in Figure~\ref{fig:overview}.

	Formally, consider a stage $g_i$ of the $n$-stage pre-trained vision model $G=\{g_i\}^{n}_{i=1}$ and an input image $\boldsymbol{\mathrm{x}}^{j}$ on a downstream dataset $\mathcal{D}=\left\{\left(\boldsymbol{\mathrm{x}}^{j}, \boldsymbol{\mathrm{y}}^{j}\right)\right\}_{j=1}^{n_{d}}$, we define the multi-stage intermediate feature maps of $\boldsymbol{\mathrm{x}}^{j}$ as $\{\boldsymbol{\mathrm{x}}^{j}_1, \boldsymbol{\mathrm{x}}^{j}_2, \cdots, \boldsymbol{\mathrm{x}}^{j}_{n}\}$, thus the original network output $\boldsymbol{\mathrm{x}}^{j}_{i}$ after the stage $g_i$ can be represented as $\boldsymbol{\mathrm{x}}^{j}_{i}=g_i(\boldsymbol{\mathrm{x}}^{j}_{i-1})$. Suppose that each prompt block after stage $g_i$ is represented by a function $f_i$, this prompt-based blending representation $\widetilde{\mathrm{x}}^{j}_i$, with the intermediate feature map $x_i^j$ and the learned task-specific prompt $f_i(x_{i}^{j})$ aggregated, can be written as
	\begin{equation}\label{eq:protuning}
		\widetilde{\mathrm{x}}^{j}_i=x_{i}^{j}+\beta \cdot f_i(x_{i}^{j}),
	\end{equation}
	where $\beta$ denotes a learnable parameter to balance the two terms. In this way, the task-specific knowledge of downstream data can be sufficiently distilled from different level representations. Note that, the prompt-based blending representation encourages a balance between the intermediate feature map and the learned prompt when pondering semantics of different stages, enabling adaptive control. Quite intuitively, our prompt blocks conducts a generic architectural modification to re-purpose a frozen pre-trained vision model for adapting to a downstream task via minor additional parameters. To this end, we implement each prompt block with three lightweight convolutional layers along with a SE module~\cite{hu2018squeeze}: 1$\times$1 convolution, 5$\times$5 depthwise convolution~\cite{chollet2017xception}, and 1$\times$1 convolution. Our default setting is appending one prompt block after each stage. Further analysis about the number of prompt blocks and inserting positions can be referred to Sec.~\ref{Ablation}. Finally, an additional fully-connected layer $h$ is appended to the top layer of pre-trained vision models for the network output. Details of configurations and stage partitioning can be found in the supplementary material.

	\subsection{Optimization}
	\label{section3.2}
	In our Pro-tuning, suppose the pre-trained vision model $G^{\phi_{0}}=\{g^{\phi_{0}}_{i}\}_{i=1}^{n}$ is parameterized by $\phi_{0}$, we define that all additional modules are parameterized by $\theta$, including $\{f^{\theta}_{i}\}_{i=1}^{n}$ and $h^{\theta}$. During training, we only update a few parameters $\theta$ with the pre-trained parameters $\phi_{0}$ frozen. Our experiments show that updating $\phi$ and $\theta$ simultaneously hurts generalization in Sec~\ref{Ablation}, the reason could be attributed to the much fewer number of downstream images compared with the large-scale dataset used to pre-train $\phi$. Training only the parameters $\theta$ makes our Pro-tuning \textit{modular} -- it can use an off-the-shelf pre-trained vision model removing the need for modifying or re-training, merely adding a small number of additional parameters per task. It is applicable to the common scenario of cloud services, where a diverse set of downstream tasks need to be solved by a given vision model, thus this parameter sharing of frozen pre-trained models is significantly appealing. Thus, a parameter-efficient vision tuning paradigm, \textit{i.e.}, Pro-tuning, can be formulated as
	\begin{equation}\label{eq:Pro-tuning}
		\begin{split}
			&\theta^{*}=\arg \underset{\theta}{\min } \frac{1}{\left|\mathcal{D}\right|} \sum_{j=1}^{n_{d}}\ell\left(h^{\theta}(\tilde{\mathbf{x}}_{i}^{j}), \mathbf{y}^{j}\right), \\ 
			& \text{where} \ \  \tilde{\mathbf{x}}_{i}^{j}=g_{i}^{\phi_{0}}(\mathbf{x}_{i-1}^{j})+\beta f_{i}^{\theta}\left(g_{i}^{\phi_{0}}(x_{i-1}^{j})\right), \quad i = 2, \cdots, n \\
			&\quad \quad \ \; \ \  \tilde{\mathbf{x}}_{1}^{j}=g_{1}^{\phi_{0}}(\mathbf{x}^{j})+\beta f_{1}^{\theta}\left(g_{i}^{\phi_{0}}(\mathbf{x}^{j})\right).
		\end{split}
	\end{equation}
	During training, we simply minimize the prediction error using the cross-entropy~(CE) loss with the pre-trained model frozen. By only training a few additional parameters, the gradients can be back-propagated all the way through the pre-trained model, distilling the rich knowledge encoded in pre-trained parameters for learning the task-specific prompt.

	\begin{figure*}[t]
		\centering
		\includegraphics[scale=0.735]{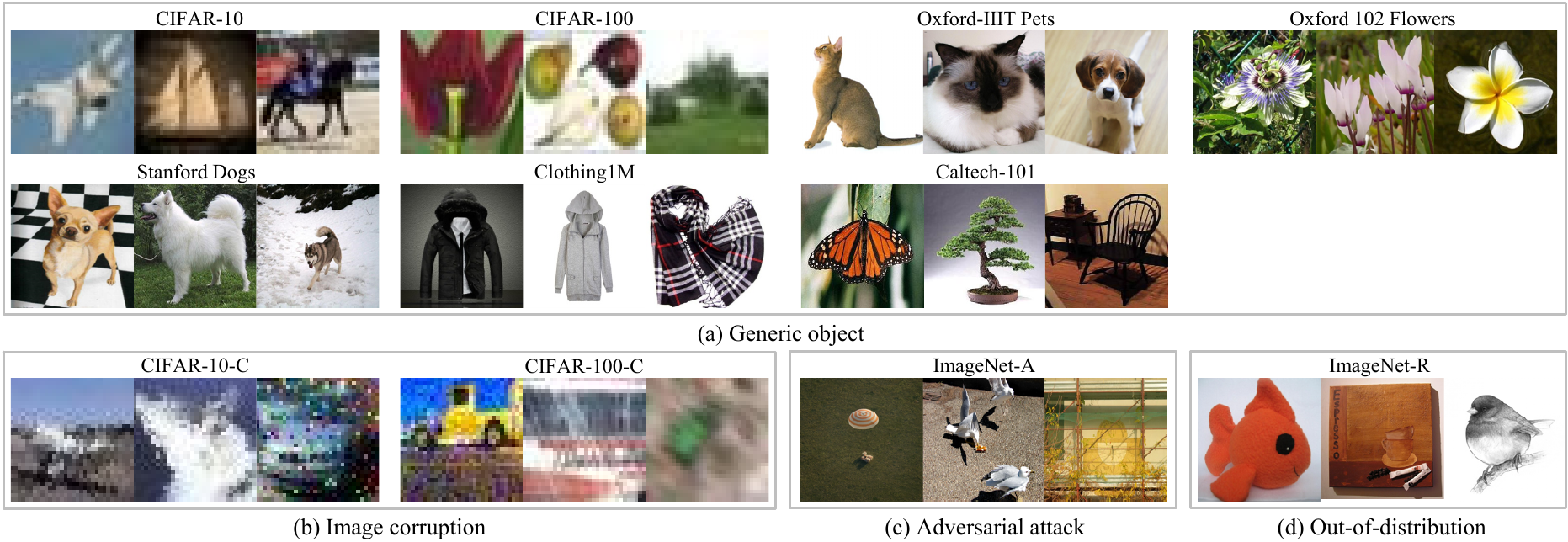} 
		\caption{Dataset examples for the used classification tasks.}
		\label{fig:dataset_example}
	\end{figure*}
	
	\begin{table*}
		\centering
		\scriptsize
		\vspace{3mm}
		\renewcommand{\arraystretch}{0.98}{
			\setlength{\tabcolsep}{0.05mm}{
				\caption{The performance of our proposed Pro-tuning and other tuning paradigms under three CNN-based pre-trained models on nine benchmarks. Notably, $\text{Params}^{\dagger}$ denotes the maximum number of trainable parameters. ``TS'': training from scratch, ``HR'': head retraining, ``FT'': fine-tuning.}\label{tab:main_results_1}
				\begin{tabular}{p{1.5cm}lp{1.1cm}<{\centering}p{0.9cm}<{\centering}p{0.9cm}<{\centering}p{0.9cm}<{\centering}p{0.9cm}<{\centering}p{0.9cm}<{\centering}p{0.9cm}<{\centering}p{0.9cm}<{\centering}p{0.9cm}<{\centering}p{0.9cm}<{\centering}p{0.9cm}<{\centering}p{0.9cm}<{\centering}}
					\toprule
					\multirow{3}{*}{Backbone} &\multirow{3}{*}{Method} & $\text{Params}^{\dagger}$ & \multicolumn{2}{c}{C-10-LT} &  \multicolumn{2}{c}{C-100-LT} & \multirow{3}{*}{C-10} & \multirow{3}{*}{C-100} & \multirow{3}{*}{PET} & \multirow{3}{*}{Flower} & \multirow{3}{*}{DOG} & \multirow{3}{*}{C1M} & \multirow{3}{*}{C-101} \\ 
					\cmidrule{3-3}\cmidrule(lr){4-5}\cmidrule(lr){6-7}
					& & (M) & IR100 & IR50 & IR100 & IR50 &  & & & & &  & \\
					\midrule
					\multicolumn{14}{l}{\emph{CNN-based}}\\
					\midrule
					\multirow{6}{*}{ResNet-50} 
					& TS~\cite{you2020co}               & 23.75 & 80.4 & 84.6 & 42.6 & 50.1 & 96.4 & 80.4 & 78.2 & 77.8 & 66.8 & 77.0 & 77.2 \\ 
					& HR~\cite{donahue2014decaf}       & 0.25 & 83.6 & 85.0 & 56.1 & 60.2 & 86.1  & 69.7 & 91.8  & 93.0 & 88.9 & 66.5 & 90.0 \\ 
					& Partial-1~\cite{yosinski2014transferable} & 4.71 & 86.5 & 88.9 & 57.9 & 62.0 & 95.5 & 79.3 & 91.2 & 96.8 & 87.4 & 75.3 & 91.5 \\
					& Partial-2~\cite{yosinski2014transferable} & 9.17 & 87.3 & 89.5 & 56.8 & 62.1 & 96.1 & 80.9 & 91.4  & 97.3 & 85.8 & 76.5 & 92.2 \\
					& FT~\cite{agrawal2014analyzing} & 23.75 & 89.7 & 92.0 & 60.5 & 65.7 & 97.5 & \textbf{84.5} & \textbf{92.5}  & \textbf{98.3} & 86.5 & \textbf{78.7} & 93.4 \\
					& Pro-tuning                                & 3.86 & \textbf{89.9} & \textbf{92.2} & \textbf{61.2} & \textbf{66.0}& \textbf{97.6} & 84.0 & \textbf{92.5} & 98.0 & \textbf{89.1} & 77.5 & \textbf{93.6}  \\

					\cmidrule{1-14}
					\multirow{6}{*}{ResNet-101} 
					& TS~\cite{you2020co}               & 42.75 & 80.4  & 84.6  & 44.6 & 50.6 & 97.1 & 81.9 & 77.7 & 78.0 & 69.0 & 77.8 & 77.4 \\ 
					& HR~\cite{donahue2014decaf}        & 0.25 & 85.3 & 86.5 & 58.9 & 62.7 & 87.6 & 71.9 & 92.4 & 92.8 & \textbf{90.5} & 67.2 & 90.7 \\ 
					& Partial-1~\cite{yosinski2014transferable} & 4.71 & 87.2 & 89.7 & 59.5 & 63.9 & 95.5 & 80.2 & 91.4 & 96.8 & 89.5 & 75.3 & 92.4 \\ 
					& Partial-2~\cite{yosinski2014transferable} & 9.17 & 87.8 & 90.5 & 59.0 & 63.7 & 96.4 & 81.7 & 91.5 & 97.6 & 88.0 & 76.3 & 92.8 \\ 
					&  FT~\cite{agrawal2014analyzing} & 42.75 & 90.8 & 92.7 & 61.8 & 67.2 & \textbf{98.2} & \textbf{87.0} & 93.7 & 98.4 & 87.9 & 78.5 & 93.7 \\ 
					& Pro-tuning                                & 3.86 & \textbf{91.0} & \textbf{93.3} & \textbf{63.9} & \textbf{68.4} & 98.0 & 86.6 & \textbf{93.9} & \textbf{98.6} & 90.3 & \textbf{79.0} & \textbf{93.8} \\

					\cmidrule{1-14}
					\multirow{6}{*}{RegNetX-32}
					& TS~\cite{you2020co}               & 105.59 & 77.5 & 83.0 & 39.2 & 46.3 & 96.8 & 81.5 & 77.8 & 78.3 &	70.2 & 77.5 & 77.2 \\  
					& HR~\cite{donahue2014decaf}        & 0.30 & 83.4 &	84.5 & 55.1 & 59.6 & 86.6 & 71.4 & 93.4 & 94.3 & \textbf{95.0} & 69.7 & 91.7 \\
					& Partial-1~\cite{yosinski2014transferable} & 3.69 & 86.2 & 88.6 & 57.6 & 62.2 & 93.9 & 78.1 & 92.6 & 97.3 & 94.2 & 76.2 & 93.7 \\
					& Partial-2~\cite{yosinski2014transferable} 		& 10.05 & 86.7 & 88.8 & 57.7 & 62.4 & 94.7 & 78.7 & 93.0 & 97.4 & 94.1 & 75.6 & 93.6 \\
					& FT~\cite{agrawal2014analyzing} 		& 105.59 & 91.0 & 92.7 & 63.7 & \textbf{68.9} & \textbf{98.4	} & 87.2 & 94.0 &\textbf{98.7} & 88.0 & \textbf{78.8} & 95.1 \\
					& Pro-tuning                                & 5.91 & \textbf{91.2} & \textbf{93.0} &	\textbf{63.9} & 68.2 & 97.8 & \textbf{87.4} & 	\textbf{94.3} & 97.9 & 94.7 & 77.9 & \textbf{95.6} \\ 
					

					\bottomrule
			\end{tabular}}
		}
		\vspace{-0.5cm}
	\end{table*}
	
	\section{Experiment}
	\label{experiment}
	In this section, we systematically demonstrate the capability of the proposed Pro-tuning. First, we briefly introduce experimental settings in Sec~\ref{Exp_setting}. Then, we perform extensive experiments to validate Pro-tuning on image classification under five scenarios, including generic objects, class imbalance, image corruption, adversarial robustness, and out-of-distribution generation in Sec~\ref{Classification}. Moreover, we evaluate Pro-tuning on dense prediction tasks such as objection detection and semantic segmentation in Sec~\ref{Dense}. Finally, we provide detailed ablation studies to analyze Pro-tuning in Sec~\ref{Ablation}. 
	
	
	\subsection{Experimental Setting}
	\label{Exp_setting}
	To exhaustively evaluate the proposed Pro-tuning, we perform detailed experiments on fifteen long-standing benchmarks. Image examples of the used classification datasets are illustrated in Figure~\ref{fig:dataset_example}. For abbreviation in Table~\ref{tab:main_results_1} and Table~\ref{tab:main_results_2}, ``C-10-LT'': long-tailed CIFAR-10, ``C-100-LT'': long-tailed CIFAR-100, ``C-10'': CIFAR-10, ``C-100'': CIFAR-100, ``PET'': Oxford-IIIT Pets, ``Flower'': Oxford Flowers-102, ``DOG'': Stanford Dogs, ``C1M'': Clothing1M, ``C-101'': Caltech-101We show that the proposed Pro-tuning facilitates a board range of vision tasks under seven advanced pre-trained models, including CNN-based and Transformer-based architectures: ResNet-50~\cite{he2016deep}, ResNet-101~\cite{he2016deep}, RegNetX-32G~\cite{radosavovic2020designing}, DeiT-S~\cite{touvron2021training}, DeiT-B~\cite{touvron2021training}, Swin-S~\cite{liu2021swin}, Swin-B~\cite{liu2021swin}. During training, we first use the pre-trained weights on ImageNet to initialize pre-trained vision models. In Table~\ref{tab:main_results_1} and Table~\ref{tab:main_results_2}, we compute the maximum number of trainable parameters to measure the parameter efficiency, denote by $\text{Params}^{\dagger}$. More experimental details can be referred to in the supplementary material.

	\begin{table*}
		\centering
		\scriptsize
		\vspace{3mm}
		\renewcommand{\arraystretch}{0.98}{
			\setlength{\tabcolsep}{0.05mm}{
				\caption{The performance of our proposed Pro-tuning and other tuning paradigms under four Transformer-based pre-trained models on nine benchmarks. Notably, $\text{Params}^{\dagger}$ denotes the maximum number of trainable parameters. ``TS'': training from scratch, ``HR'': head retraining, ``FT'': fine-tuning.}\label{tab:main_results_2}
				\begin{tabular}{p{1.5cm}lp{1.1cm}<{\centering}p{0.9cm}<{\centering}p{0.9cm}<{\centering}p{0.9cm}<{\centering}p{0.9cm}<{\centering}p{0.9cm}<{\centering}p{0.9cm}<{\centering}p{0.9cm}<{\centering}p{0.9cm}<{\centering}p{0.9cm}<{\centering}p{0.9cm}<{\centering}p{0.9cm}<{\centering}}
					\toprule
					\multirow{3}{*}{Backbone} &\multirow{3}{*}{Method} & $\text{Params}^{\dagger}$ & \multicolumn{2}{c}{C-10-LT} &  \multicolumn{2}{c}{C-100-LT} & \multirow{3}{*}{C-10} & \multirow{3}{*}{C-100} & \multirow{3}{*}{PET} & \multirow{3}{*}{Flower} & \multirow{3}{*}{DOG} & \multirow{3}{*}{C1M} & \multirow{3}{*}{C-101} \\ 
					\cmidrule{3-3}\cmidrule(lr){4-5}\cmidrule(lr){6-7}
					& & (M) & IR100 & IR50 & IR100 & IR50 &  & & & & &  & \\
					\cmidrule{1-14}
					\multicolumn{14}{l}{\emph{Transformer-based}}\\
					\cmidrule{1-14}
					\multirow{6}{*}{DeiT-S}
					& TS~\cite{you2020co}               & 21.71 & 58.9 & 67.9 & 33.5 & 37.9 & 95.3 & 74.1 & 41.8 & 68.9 & 34.4 & 68.4 & 50.1  \\ 
					& HR~\cite{donahue2014decaf}        & 0.05 & 83.6 & 86.9 & 63.1 & 66.8 & 92.6 & 77.0 & 92.8 & 89.0 & \textbf{92.7} & 66.2& 90.6 \\
					& Partial-1~\cite{yosinski2014transferable} 		& 1.23 & 87.7 & 89.2 & 62.2 & 66.5 & 95.6 & 80.9 & 93.6 & 92.7 & 92.6 & 70.9 & 91.5 \\
					& Partial-2~\cite{yosinski2014transferable} 		& 1.82 & 89.3 & 91.6 & 64.6 & 68.9 & 96.9 & 83.5 & 94.2 & 95.6 & 92.2 & 74.6 & 92.7 \\
					& FT~\cite{agrawal2014analyzing} 		& 21.71 & 92.0 & 94.0 & 69.4 & 73.6 & \textbf{98.5} & 87.5 & 93.8 & \textbf{98.3} & 87.5 & \textbf{78.5} & \textbf{94.8} \\
					& Pro-tuning                                & 0.85 &\textbf{92.6} & \textbf{94.2} & \textbf{70.0} & \textbf{74.4} & 98.2 & \textbf{88.0} & \textbf{94.0} & 97.0 & 92.4 & 78.3 & 94.3 \\ 
					
					

					\cmidrule{1-14}
					\multirow{6}{*}{Swin-S}
					& TS~\cite{you2020co}               & 48.93 & 62.6 & 70.8 & 40.7 & 47.2	& 95.1 & 79.9 & 52.9 & 74.3 & 45.9 & 21.6 & 66.1 \\
					& HR~\cite{donahue2014decaf}        & 0.09 & 83.5 & 87.6 & 61.9 & 66.4 & 93.3 & 78.4 & 93.3 & 88.8 & \textbf{95.0} & 66.5 & 91.6 \\
					& Partial-1~\cite{yosinski2014transferable}		& 2.45 & 86.6 & 89.7 & 64.1 & 69.0 & 94.4 & 80.8 & 93.5 & 93.3 & 94.9 & 70.2 & 92.5 \\
					& Partial-2~\cite{yosinski2014transferable} 		& 4.82 & 90.2 & 92.2 & 65.3 & 70.2 & 96.8 & 84.6 & 94.9 & 95.9 & 94.8 & 76.0 & 93.5 \\
					&
					FT~\cite{agrawal2014analyzing} 		& 48.93 & 93.6 & \textbf{95.8} & 70.6 & \textbf{75.5} & \textbf{99.0} & \textbf{89.4	} & 94.6 & \textbf{98.9} & 90.2 & \textbf{80.0} & \textbf{96.2} \\
					& Pro-tuning                                & 1.54 & \textbf{93.8} & 95.5 & \textbf{71.2} & 75.3 & 98.6 & 88.7 & \textbf{95.0} & 97.0 & 94.5 & 78.4 & 94.7 \\
					

					\cmidrule{1-14}
					\multirow{6}{*}{DeiT-B}
					& TS~\cite{you2020co}               & 85.89  & 58.2  & 65.8 & 31.3 & 35.2& 95.5 & 72.5 & 40.2 & 71.1 & 29.8 & 58.1 & 50.3 \\
					& HR~\cite{donahue2014decaf}        & 0.09 & 87.5 & 90.3 & 67.3 & 71.4 & 95.1 & 81.5 & 93.6 & 93.0 & \textbf{96.0} & 69.9 & 91.8 \\
					& Partial-1~\cite{yosinski2014transferable} & 4.82 & 89.8 & 92.0 & 65.8 & 70.5 & 97.2 & 84.2 & 94.4 & 95.1 & 95.8 & 72.2 & 93.2 \\
					& Partial-2~\cite{yosinski2014transferable} & 7.18 & 91.7 & 93.7 & 69.2 & 73.8 & 98.1 & 87.2 & 94.8 & 96.8 & 95.6 & 76.1 & 93.9 \\
					& FT~\cite{agrawal2014analyzing}& 85.89 & \textbf{93.8} & 94.8 & 73.1 & 76.5 & 98.7 & 88.9 & 94.0 & \textbf{98.5} & 91.2 & \textbf{78.7} & \textbf{95.9} \\
					& Pro-tuning                                & 3.17 & 93.5 & \textbf{95.5} & \textbf{74.0} & \textbf{77.7} & \textbf{98.8} & \textbf{89.7} & \textbf{94.7} & 97.7 & 95.9 & 78.5 & 95.6 \\
					

					\cmidrule{1-14}
					\multirow{6}{*}{Swin-B}
					& TS~\cite{you2020co}               & 86.87 & 62.3 & 70.1 & 40.9 & 46.9& 94.8 & 79.5 & 44.8 & 76.7 & 5.2 & 19.6 & 67.4 \\
					& HR~\cite{donahue2014decaf}        & 0.12 & 84.6 & 88.2 & 63.0 &67.5 & 94.0 & 79.1 & 92.8 & 90.6 & \textbf{95.6} & 68.2 & 92.1 \\
					& Partial-1~\cite{yosinski2014transferable} & 4.32 & 86.5 & 89.4 & 64.3	 & 68.7 & 94.8 & 81.0 & 93.6 & 94.2 & 95.5 & 71.4 & 92.5 \\
					& Partial-2~\cite{yosinski2014transferable} & 8.52 & 88.7 & 91.4 & 64.4	 & 69.6 & 96.9 & 84.3 & 94.4 & 95.4 & 95.5 & 95.7 & 93.4 \\
					& FT~\cite{agrawal2014analyzing}& 86.87 & 93.9 & 95.3 & \textbf{71.6}	 & \textbf{76.2} & 98.5 & \textbf{89.4} & \textbf{95.1} & \textbf{98.9} & 91.8 & \textbf{79.9} & \textbf{96.5} \\
					& Pro-tuning                                & 2.67 & \textbf{94.2} & \textbf{95.4} & 71.2 &75.7 & \textbf{98.6} & 88.8 & 95.0 & 96.7 & 95.5 & 77.7 & 93.8 \\

					\bottomrule
			\end{tabular}}
		}
		\vspace{-0.5cm}
	\end{table*}

	To put our results in perspective, we compare the proposed Pro-tuning with a series of commonly used protocols, \textit{i.e.}, 1) \textit{fine-tuning}~\cite{agrawal2014analyzing}, which fine-tunes the weight of the pre-trained vision model while retraining the new head classifier on downstream tasks; 2) \textit{partial fine-tuning}~\cite{yosinski2014transferable}, which fine-tunes the last several layers of the pre-trained vision model while freezing the remaining parts and retraining a new head classifier on downstream tasks, including two variants \textit{Partial-1} and \textit{Partial-2} according to the number of trainable parameters; 3) \textit{head retraining}~\cite{donahue2014decaf}, which retrains the head classifier while freezing the weight of the pre-trained model; 4) \textit{training from scratch}, which trains the whole vision model from random initialization.

	\subsection{Image Classification}
	\label{Classification}
	
	\noindent \textbf{Class Imbalance Transferability.}\,\,~We conduct extensive experiments on the long-tailed CIFAR-10 and CIFAR-100 datasets. Table~\ref{tab:main_results_1} and Table~\ref{tab:main_results_2} show the results of Pro-tuning and other tuning paradigms. Under three pre-trained CNNs, Pro-tuning outperforms other tuning methods including fine-tuning~(FT) and the most lightweight tuning method head retraining~(HR) in most scenarios. Specifically, for long-tailed CIFAR-100 with imbalanced ratio 100, Pro-tuning achieves superior validation accuracy~(63.9\%) compared with fine-tuning~(61.8\%) and head retraining~(58.9\%) under ResNet-101~\cite{he2016deep}. Regarding long-tailed CIFAR-10 with imbalanced ratio 50, Pro-tuning brings in 0.6\% absolute accuracy improvement and reduces the trainable parameters by 91.0\% compared with fine-tuning under ResNet-101~\cite{he2016deep}. Under the larger-capacity RegNetX-32~\cite{radosavovic2020designing}, Pro-tuning also achieves superior performance~(91.2\% \textit{vs.} 91.0\%) over fine-tuning with a large reduction of trainable parameters by 94.4\% on long-tailed CIFAR-10 with imbalanced ratio 100. For state-of-the-art transformer-based architectures, Pro-tuning also exhibits excellent performance. In particular, Pro-tuning consistently outperforms fine-tuning under DeiT-S~\cite{touvron2021training} on various imbalanced ratios, with only 25.5x fewer trainable parameters. On long-tailed CIFAR-100 with imbalanced ratio 50, Pro-tuning brings in a gain of 1.2\% validation accuracy over fine-tuning under DeiT-B~\cite{touvron2021training} while reducing the trainable parameters by 96.3\%. Moreover, compared with fine-tuning on long-tailed CIFAR-10 with imbalanced ratio 100 under Swin-S~\cite{liu2021swin}, Pro-tuning achieves superior validation accuracy using 31.8x fewer trainable parameters. Regarding Swin-B~\cite{liu2021swin}, Pro-tuning surpasses fine-tuning on long-tailed CIFAR-10 with imbalanced ratio 50 and 100 with a large reduction of trainable parameters by 96.2\%.

	\vspace{3pt}
	\noindent \textbf{Generic Object Transferability.}\,\,~To reveal the transferability of generic objects, we report the quantitative results on seven downstream datasets with generic objects. As shown in Table~\ref{tab:main_results_1} and Table~\ref{tab:main_results_2}, Pro-tuning shows remarkable performance under a series of pre-trained models. Specifically, Pro-tuning reaches the identical validation accuracy of fine-tuning under RegNetX-32~\cite{radosavovic2020designing} on Oxford-IIIT Pets, while only using 17.9x fewer trainable parameters. Under three pre-trained CNNs, Pro-tuning brings in a gain of 9.8\%$\sim$11.5\% validation accuracy over head retraining on CIFAR-10 while obtaining comparable performances to fine-tuning. Moreover, Pro-tuning achieves superior validation accuracy with 11.1x trainable parameters over fine-tuning under ResNet-101~\cite{he2016deep} on Caltech-101. For advanced transformer-based architectures, Pro-tuning achieves quite competitive performance compared with fine-tuning and head retraining. Specifically, Pro-tuning brings in 1.0\% absolute accuracy improvement over fine-tuning with 27.1x fewer trainable parameters under DeiT-B on CIFAR-100. Moreover, Pro-tuning obtains superior performance~(95.0\% \textit{vs.} 94.6\%) than fine-tuning while reducing the trainable parameters by 96.9\% under Swin-S~\cite{liu2021swin} on Oxford-IIIT Pets. Under Swin-B~\cite{liu2021swin}, pro-tuning brings in a gain of 4.6\% validation accuracy over head retraining and achieves comparable performance to fine-tuning with 32.5x fewer trainable parameters on CIFAR-10. 
	
	\vspace{3pt}
	\noindent \textbf{Image Corruption Robustness.}\,\,~We show the results on CIFAR-10-C and CIFAR-100-C in Table~\ref{tab:corruption}. Notably, our proposed Pro-tuning consistently performs better than all the other methods on two datasets under three pre-trained models. On CIFAR-10-C, our Pro-tuning brings in a gain of +0.4\%$\sim$2.7\% over fine-tuning, with only 6.5x$\sim$33.5x fewer trainable parameters. Compared with head retraining, Pro-tuning achieves better transfer performance with at least +10.5\% absolute validation accuracy gains. On CIFAR-100-C, Pro-tuning brings +1.2\% absolute validation accuracy gains than fine-tuning with more than 83.9\% drop in trainable parameters (3.823M \textit{vs.} 23.714M) under ResNet-50. The dramatic improvements demonstrate the capability of our proposed Pro-tuning to defend against image corruption.
	
	\begin{table*}[h]
		\centering
		\scriptsize
		\vspace{3mm}
		\renewcommand{\arraystretch}{1.0}{
			\setlength{\tabcolsep}{2.0mm}{
				\caption{The performance of Pro-tuning and other tuning paradigms on CIFAR-10-C and CIFAR-100-C. ``FT'': fine-tuning, ``HR'': head retraining, ``TS'': training from random initialization.}\label{tab:corruption}
				\begin{tabular}{p{2cm}lp{2cm}<{\centering}p{2cm}<{\centering}p{2cm}<{\centering}p{2cm}<{\centering}}
					\toprule
					\multirow{3}{*}{Backbone} &\multirow{3}{*}{Method}  & \multicolumn{2}{c}{ CIFAR-10-C} &  \multicolumn{2}{c}{ CIFAR-100-C}  \\ 
					\cmidrule(lr){3-4} \cmidrule(lr){5-6}
					& & Params~(M) & Top1 Acc.~(\%) & Params~(M) & Top1 Acc.~(\%)  \\
					\midrule
					\multirow{4}{*}{ResNet-50} 
					& FT~\cite{agrawal2014analyzing}    & 23.532 & 80.8 & 23.714 & 57.4  \\
					& HR~\cite{donahue2014decaf}        & 0.021 & 62.2 & 0.210 & 43.6  \\ 
					& TS~\cite{you2020co}               & 23.532 & 83.0 & 23.714 & 58.2  \\ 
					& Pro-tuning                        & 3.634 & \textbf{83.5} & 3.823 & \textbf{58.6} \\

					\cmidrule{1-6}
					\multirow{4}{*}{DeiT-S} 
					& FT~\cite{agrawal2014analyzing}    & 21.670  & 91.1  & 21.704 & 72.8 \\ 
					& HR~\cite{donahue2014decaf}         & 0.004  &  79.7 & 0.040  & 59.2  \\ 
					& TS~\cite{you2020co}                & 21.670  &  85.9 & 21.704 & 59.1  \\ 
					& Pro-tuning                         & 0.773   & \textbf{91.5}  & 0.808  & \textbf{73.0} \\

					\cmidrule{1-6}
					\multirow{4}{*}{Swin-S}
					& FT~\cite{agrawal2014analyzing} 	 & 48.845 & 88.2 & 48.914 & 70.4  \\
					& HR~\cite{donahue2014decaf}         & 0.008 & 79.7 & 0.080 & 59.1  \\
					& TS~\cite{you2020co}                & 48.845 & 82.7 & 48.914 & 59.1  \\  
					& Pro-tuning                         & 1.458  & \textbf{90.2} & 1.527 &  \textbf{70.7} \\ 
					
					\bottomrule
			\end{tabular}}
		}
		\vspace{-0.5cm}
	\end{table*}
	
	\vspace{3pt}
	\noindent \textbf{Adversarial Robustness.}\,\,~We employ ImageNet-A to evaluate all the tuning methods on adversarial examples in Table~\ref{tab:INRINA}. The performances of all tuning methods are relatively low on ImageNet-A, as this task is quite difficult. Even though, our Pro-tuning also outperforms other tuning methods under four advanced pre-trained models. For example, Pro-tuning surpasses fine-tuning with bringing in +4.8\% absolute performance gains under DeiT-S, while requires only 25.7x fewer trainable parameters. Moreover, Pro-tuning brings in significant accuracy gains of +6.9\% over fine-tuning with the reduction of trainable parameters by 96.7\% under Swin-S.
	
	\begin{table*}[h]
		\centering
		\scriptsize
		\vspace{3mm}
		\renewcommand{\arraystretch}{0.9}{
			\setlength{\tabcolsep}{2.0mm}{
				\caption{The performance of Pro-tuning and other tuning paradigms on ImageNet-A and ImageNet-R. ``FT'': fine-tuning, ``HR'': head retraining, ``TS'': training from random initialization.}\label{tab:INRINA}
				\begin{tabular}{p{2cm}lp{2cm}<{\centering}p{2cm}<{\centering}p{2cm}<{\centering}p{2cm}<{\centering}}
					\toprule
					\multirow{3}{*}{Backbone} &\multirow{3}{*}{Method}  & \multicolumn{2}{c}{ImageNet-A} &  \multicolumn{2}{c}{ImageNet-R}  \\ 
					\cmidrule(lr){3-4} \cmidrule(lr){5-6}
					& & Params~(M) & Top1 Acc.~(\%) & Params~(M) & Top1 Acc.~(\%)  \\
					\midrule
					\multirow{4}{*}{DeiT-S} 
					& FT~\cite{agrawal2014analyzing}    & 21.743 & 15.6 & 21.743 & 40.6 \\ 
					& HR~\cite{donahue2014decaf}        & 0.081 & 19.2 & 0.081 & 40.9   \\ 
					& TS~\cite{you2020co}               & 21.743 & 6.5 & 21.743 & 24.9  \\ 
					& Pro-tuning                        & 0.846  &   \textbf{20.4} & 0.846 & \textbf{41.6} \\  

					\cmidrule{1-6}
					\multirow{4}{*}{Swin-S} 
					& FT~\cite{agrawal2014analyzing} & 48.991 &  26.5 & 48.991 & 45.2 \\   
					& HR~\cite{donahue2014decaf}         & 0.154 &  32.8 & 0.154 & 45.2  \\ 
					& TS~\cite{you2020co}                & 48.991  &  12.6 & 48.991 &  30.5 \\   
					& Pro-tuning                         & 1.604 &   \textbf{33.4} & 1.604 & \textbf{46.4} \\ 

					\cmidrule{1-6}
					\multirow{4}{*}{DeiT-B}
					& FT~\cite{agrawal2014analyzing} 	 & 85.952 & 24.1 & 85.952 & \textbf{44.9}  \\    
					& HR~\cite{donahue2014decaf}         & 0.154  &  27.8 & 0.154  &  44.0  \\    
					& TS~\cite{you2020co}                & 85.952 & 5.9 & 85.952 &  23.6 \\   
					& Pro-tuning                         & 3.236  & \textbf{28.5}  & 3.236  & 44.3 \\   
					
					\cmidrule{1-6}
					\multirow{4}{*}{Swin-B}
					& FT~\cite{agrawal2014analyzing} 	 & 86.948 & 33.6  & 86.948 &  \textbf{47.4} \\ 
					& HR~\cite{donahue2014decaf}         & 0.205  &  36.5 & 0.205  & 46.6 \\   
					& TS~\cite{you2020co}                & 86.948 & 13.7 & 86.948 &  30.3 \\  
					& Pro-tuning                         & 2.749  & \textbf{37.7} & 2.749  &   46.6 \\  
					\bottomrule
			\end{tabular}}
		}
		\vspace{-0.5cm}
	\end{table*}
	
	\vspace{3pt}
	\noindent \textbf{Out-of-distribution Generalization.}\,\,~We present the validation accuracy and trainable parameters on ImageNet-R in Table~\ref{tab:INRINA}. Note that the proposed Pro-tuning consistently surpasses head retraining under four pre-trained models. Specifically, Pro-tuning brings in a gain of +1.0\% validation accuracy over fine-tuning using fewer trainable parameters under DeiT-S. Regarding the larger-capacity Swin-S, Pro-tuning obtains 46.2\% validation accuracy, +1.2\% higher than fine-tuning and head retraining.

	\subsection{Dense Prediction}
	\label{Dense}
	
	\vspace{3pt}
	\noindent \textbf{Object detection.}\,\,We evaluate the proposed Pro-tuning over the object detection task under Cascade Mask RCNN~\cite{cai2019cascade} and HTC~\cite{chen2019hybrid} with ImageNet pre-trained ResNet-50. As shown in Table~\ref{object}, one can observe that our Pro-tuning achieves the comparable performance with the significant reduction of trainable parameters. 
	
	\begin{table}[htp]
		\vspace{-0em}
		\small 
		\centering
		\caption{Results of object detection on COCO2017 val set under Cascade Mask RCNN~\cite{cai2019cascade} and HTC~\cite{chen2019hybrid} with ImageNet pre-training.}\label{object}
		\renewcommand{\arraystretch}{0.8}
		\setlength{\tabcolsep}{4pt}{
			\begin{tabular}{l|c|c|c|c|c|c|c|c}
				\toprule[1.5pt]
				Methods & Params & Epochs & AP & AP$_{50}$ & AP$_{75}$ & AP$_{S}$ & AP$_{M}$ & AP$_{L}$ \\
				\midrule
				\multicolumn{9}{l}{\emph{Cascade Mask RCNN}}\\
				\midrule
				Head retraining   & 53.8  & 12  & 34.2 & 52.9 & 36.8 & 19.3 & 36.9 & 45.7 \\
				Fine-tuning         & 77.1  & 12  & \textbf{41.2} & 59.4 & \textbf{45.0} & 23.9 & \textbf{44.2} & 54.4 \\
				Pro-tuning~(ours) & 57.4  & 12  & 41.1 & \textbf{60.6} & 44.9 & \textbf{24.3} & 43.5 & \textbf{55.5} \\
				\midrule
				\multicolumn{9}{l}{\emph{HTC}}\\
				\midrule
				Head retraining   & 56.8  & 12  & 37.5 & 56.3 & 40.2 & 21.1 & 40.3 & 50.3 \\
				Fine-tuning         & 80.0  & 12  & \textbf{42.3} & 61.1 & 45.8 & 23.7 & \textbf{45.6} & 56.3 \\
				Pro-tuning~(ours) & 60.4  & 12  & 42.2 & \textbf{62.3} & \textbf{46.0} & \textbf{25.2} & 45.2 & \textbf{56.7} \\
				\bottomrule[1pt]
			\end{tabular}
		}
	\end{table}
	\begin{wraptable}{r}{0.43\textwidth}
		\vspace{-1.2em}
		\small
		\centering
		\caption{Results of semantic segmentation on ADE20K val set under UperNet~\cite{xiao2018unified} and FCN~\cite{long2015fully} with ImageNet pre-training.}\label{tab:seg}
		\vspace{-0.5pt}
		\renewcommand{\arraystretch}{0.8}
		\setlength{\tabcolsep}{4.5pt}{
			\begin{tabular}{l|c|c|c}
				\toprule[1.5pt]
				Backbone      & Params & Steps & mIoU \\
				\midrule
				\multicolumn{4}{l}{\emph{FCN}}\\
				\midrule
				Head retraining   & 26.1M & 80k & 25.8\% \\
				Fine-tuning         & 49.6M & 80k & 35.9\% \\
				Pro-tuning~(ours) & 26.7M & 80k & \textbf{36.2}\% \\
				\midrule
				\multicolumn{4}{l}{\emph{UperNet}}\\
				\midrule
				Head retraining   & 43.0M & 80k & 38.7\% \\
				Fine-tuning         & 66.5M & 80k & 40.7\% \\
				Pro-tuning~(ours) & 46.6M & 80k & \textbf{41.3}\% \\
				\bottomrule[1pt]
			\end{tabular}
		}
		\vspace{-1.0em}
	\end{wraptable}
	\vspace{3pt}
	\noindent \textbf{Semantic segmentation.}\,\,To evaluate the semantic segmentation performance, we evaluate Pro-tuning under UperNet~\cite{xiao2018unified} and FCN~\cite{long2015fully} with ImageNet pre-trained ResNet-50. The results are shown in Table~\ref{tab:seg}. One can observe that our Pro-tuning consistently outperform fine-tuning and head retraining.

	\subsection{Ablation Study}
	\label{Ablation}

	In this subsection, we conduct extensive ablation studies to systematically analyze the developed Pro-tuning with the ImageNet pre-trained ResNet-50~\cite{he2016deep} and DeiT-B~\cite{touvron2021training}. 
	
	\begin{figure*}[t]
		\centering
		\includegraphics[scale=0.325]{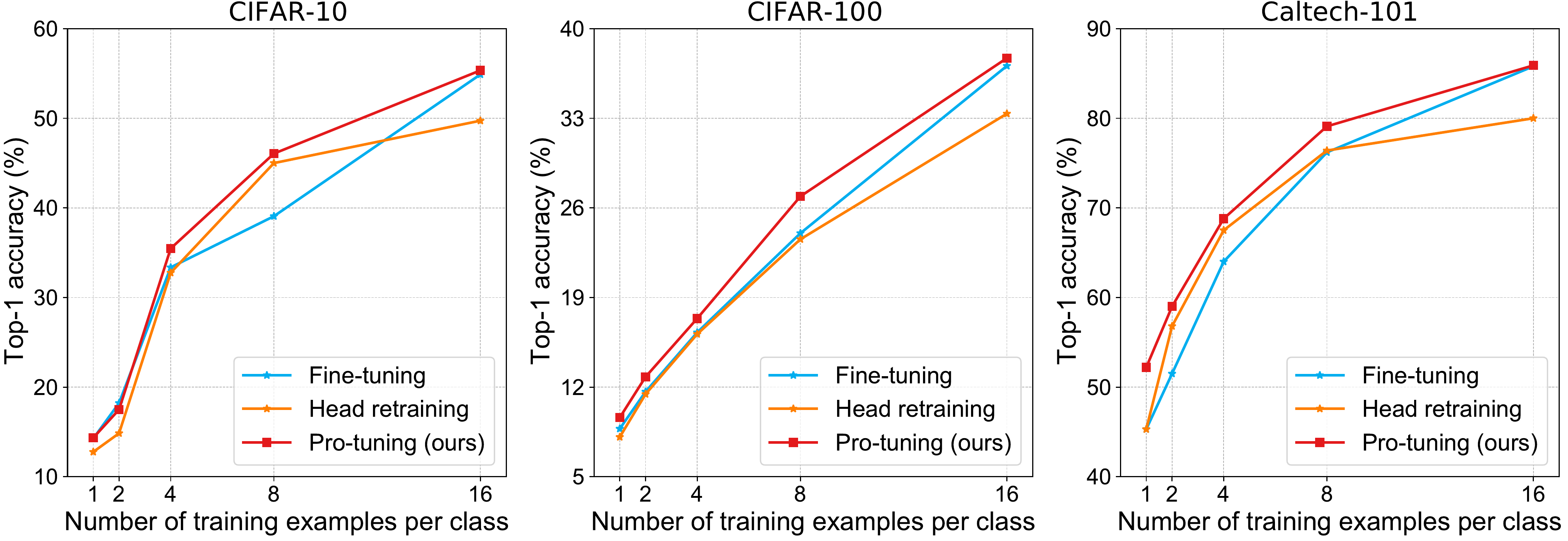} 
		\caption{Results of few-shot learning under the ImageNet pre-trained ResNet-50 on the CIFAR-10, CIFAR-100, and Caltech-101 benchmarks.}
		\label{fig:fewshot}
	\end{figure*}
	
	\vspace{3pt}
	\noindent \textbf{Analysis of few-shot learning.}\,\,To evaluate the few-shot performance, we perform few-shot transfer by reducing the number of training examples on CIFAR-10, CIFAR-100, and Caltech-101 under ResNet-50. Following the representative few-shot protocol~\cite{gao2021clip, radford2021learning, zhou2021learning}, we adopt 1, 2, 4, 8, and 16 shots for training and the full test sets for testing in Figure~\ref{fig:fewshot}. One can observe that our Pro-tuning is a strong few-shot learner, requiring only 4 shots to obtain a significant margin with +2.1\% absolute accuracy improvement over fine-tuning on CIFAR-10. Pro-tuning also achieves a superior result over head retraining and fine-tuning with 8 shots, which brings in +3.4\% and +2.9\% validation accuracy improvements on CIFAR-100. Moreover, Pro-tuning brings in a gain of 5.9\% validation accuracy over head retraining with 16 shots on Caltech-101.

	\vspace{3pt}
	\noindent \textbf{Integration with fine-tuning.}\,\,~To reveal the influence of different tuning settings, we compare Pro-tuning with one important baseline, denoted by $\text{Pro-tuning}_\text{fine-tune}$, in which our prompt blocks are trained together with the pre-trained vision model. As listed in Table~\ref{tab:ablation2}, joint training of prompt blocks and the pre-trained vision model degrades performance compared with merely updating prompt blocks, closely to fine-tuning. This is reasonable since training the pre-trained model on downstream datasets may hurt its capability of general-purpose visual modeling, due to the much less number of downstream images compared with the large-scale dataset used for pre-training, where the similar phenomena also occur in the vision-language model~\cite{jia2021scaling, tsimpoukelli2021multimodal}.

	\begin{table}[htp]
		\vspace{-0em}
		\small 
		\centering
		\caption{Validation accuracy comparisons of Pro-tuning and other tuning settings on long-tailed CIFAR-100 with imbalanced ratio 100 under ResNet-50.}\label{tab:ablation2}
		\renewcommand{\arraystretch}{1.0}
		\setlength{\tabcolsep}{4pt}{
			\begin{tabular}{l|c|c|c|c|c}
				\toprule[1.5pt]
				Methods & Training from scratch & Head retraining & Fine-tuning & Pro-tuning & $\text{Pro-tuning}_\text{fine-tune}$ \\
				\midrule
				Params~(M) & 23.71 & 0.20  & 23.71 & 3.82 & 27.33 \\
				\midrule
				Top-1 Acc.~(\%) & 42.6 & 56.1  & 60.5 & 61.2 & 60.4 \\
				\bottomrule[1pt]
			\end{tabular}
		}
	\end{table}

	\begin{wraptable}{r}{0.5\textwidth}
		\vspace{0em}
		\small 
		\centering
		\vspace{-1.3em}
		\caption{Validation accuracy comparisons of different numbers of prompt blocks per stage on long-tailed CIFAR-100 with imbalanced ratio 100 under ResNet-50.}\label{tab:ablation4}
		\vspace{-0.2em}
		\renewcommand{\arraystretch}{1.0}
		\setlength{\tabcolsep}{2.9pt}{
			\begin{tabular}{l|c|c|c}
				\toprule[1.5pt]
				Dataset & \#Block & \#Params & Top-1 Acc.  \\ 
				\midrule
				\multirow{3}{*}{CIFAR-100} & 1 & 3.82 & 83.6 \\ 
				& 2 & 7.43 & 83.5 \\ 
				& 3 & 11.05 & 83.7 \\ 
				\midrule
				\multirow{3}{*}{CIFAR-100-LT-100} & 1 & 3.82 & 61.2 \\ 
				& 2 & 7.43 & 61.3 \\ 
				& 3 & 11.05 & 61.3 \\ 
				\bottomrule[1pt]
			\end{tabular}
		}
		\vspace{-1.0em}
	\end{wraptable}
	\vspace{3pt}
	\noindent \textbf{Impact of the capacity of prompt blocks.}\,\,~To investigate the influence of the capacity of prompt blocks, we compare Pro-tuning with its variants that use two prompt blocks or three prompt blocks on CIFAR-100 and long-tailed CIFAR-100 with imbalanced ratio 100 under ResNet-50. As listed in Table~\ref{tab:ablation4}, we can observe that the increase of prompt blocks brings little help to transferability. This adequately validates that our Pro-tuning suffices to bring a significant performance gain only using one prompt block comprised of three convolutional layers. More results and analysis are provided in the supplementary material. 
	
	\vspace{3pt}
	\noindent \textbf{Influence of inserting positions.}\,\,To evaluate the impact of inserting positions of prompt blocks, we model Pro-tuning with different inserting positions and numbers under DeiT-B. As listed in Table~\ref{tab:ablation3}, we find that our adopted multi-stage inserting way achieves the best performance. Besides, it can be observed that only inserting after the first layer and our used uniformly inserting can obtain better generalization performance compared with only inserting after the last layer. A considerable reason is that low-level and mid-level semantics could be more significant than high-level semantics for transferability, as pointed in~\cite{islam2021broad, yosinski2014transferable, zhao2020makes}. More experimental results about inserting positions of prompt blocks are provided in the supplementary material. 
	
	\begin{table}[htp]
		\small 
		\centering
		\caption{Influence of inserting position for prompt blocks on long-tailed CIFAR-100 with imbalanced ratio 100 under DeiT-B. ``F1'': inserting a prompt block after the first transformer layer. ``F5'': inserting five prompt blocks after the first transformer layer. ``L1'': inserting a prompt block after the last transformer layer. ``L5'': inserting five prompt blocks after the last transformer layer. ``U5'': uniformly inserting four prompt blocks in 12 transformer layers along with one prompt block after the embedding layer, as our default setting.}\label{tab:ablation3}
		\vspace{1em}
		\renewcommand{\arraystretch}{1.0}
		\setlength{\tabcolsep}{8pt}{
			\begin{tabular}{l|c|c|c|c|c}
				\toprule[1.5pt]
				Position & F1 & F5 & L1 & L5 & U5 \\
				\midrule
				Params~(M) & 0.69 & 3.17 & 0.69 & 3.17 & 3.17 \\
				\midrule
				Top-1 Acc.~(\%) & 73.0 & 73.5 & 67.4 & 67.5 & 74.0 \\
				\bottomrule[1pt]
			\end{tabular}
		}
		\vspace{-2em}
	\end{table}

	\vspace{3pt}
	\noindent \textbf{Effect of adaptive prompt.}\,\,~To evaluate the effect of adaptive prompt, we compare Pro-tuning under different settings for the ratio $\beta$ in prompt blocks in Eq.~\eqref{eq:protuning} on long-tailed CIFAR-100 with imbalanced ratio 100 under ResNet-50. Specifically, we set $\beta$ as a fixed constant or a learnable parameter. As listed in Table~\ref{tab:ablation5}, one can observe that the learnable $\beta$ achieves superior performance over all constants. This adequately validates the effectiveness of our adaptive prompt. Interestingly, we can find that the decreasing number of $\beta$ consistently boosts performance varying from 10 to 0.1. This phenomenon implies that encouraging the feature representations from the pre-trained model may facilitate transferability when training Pro-tuning.

	\begin{table}[htp]
		\vspace{-0em}
		\small 
		\centering
		\vspace{-1em}
		\caption{Validation accuracy comparisons of different settings of $\beta$ in prompt blocks in Eq.~\eqref{eq:protuning} on long-tailed CIFAR-100 with imbalanced ratio 100 under ResNet-50.}\label{tab:ablation5}
		\renewcommand{\arraystretch}{1.0}
		\setlength{\tabcolsep}{8pt}{
			\begin{tabular}{c|c|c|c|c|c|c|c|c}
				\toprule[1.5pt]
				$\beta$ & 10 & 4 & 2 & 1 & 0.5 & 0.25 & 0.1 & Lea. \\
				\midrule
				Top-1 Acc. & 50.3 & 49.7  & 49.5 & 51.8 & 53.2 & 54.4 & 56.6 & 58.1 \\
				\bottomrule[1pt]
			\end{tabular}
		}
		\vspace{-1em}
	\end{table}
	
	\begin{wraptable}{r}{0.5\textwidth}
		\vspace{-1.3em}
		\small 
		\centering
		\caption{Validation accuracy comparisons of different kernel sizes in prompt blocks on long-tailed CIFAR-100 with imbalanced ratio 100 under ResNet-50.}\label{tab:ablation6}
		\vspace{-0.4em}
		\renewcommand{\arraystretch}{1.0}
		\setlength{\tabcolsep}{2.9pt}{
			\begin{tabular}{l|c|c|c}
				\toprule[1.5pt]
				Dataset & \#Kernel & \#Params & Top-1 Acc.  \\ 
				\midrule
				\multirow{3}{*}{CIFAR-100} & 3 & 3.76 & 83.2 \\ 
				& 5 & 3.82 & 83.6 \\ 
				& 7 & 3.91 & 83.4 \\ 
				\midrule
				\multirow{3}{*}{CIFAR-100-LT-100} & 3 & 3.67 & 60.9 \\ 
				& 5 & 3.82 & 61.2 \\ 
				& 7 & 3.91 & 61.1 \\ 
				\bottomrule[1pt]
			\end{tabular}
		}
		\vspace{-1.0em}
	\end{wraptable}

	\vspace{3pt}
	\noindent \textbf{Impact of different kernel sizes in prompt blocks.}\,\,~We evaluate Pro-tuning under the different kernel sizes $k$ of the second convolution in prompt blocks on CIFAR-100 and long-tailed CIFAR-100 with imbalanced ratio 100 under ResNet-50. The results in Table~\ref{tab:ablation6} indicate that our Pro-tuning performs outstanding transferability under diverse kernel sizes. Additionally, the performance generally improves with the increase of kernel size, then the performance will saturate when the kernel size $k=5$ is satisfied. A considerable reason is that the increase of receptive filed size typically enhances the representation ability. Nevertheless, a quite large kernel size results in difficult optimization, hence there is little benefit from a larger kernel size. Thus, we set the kernel size as 5 in our experiments unless specified otherwise.

	\section{Discussion}
	\label{discussion}
	
	In our view, the significant performance gain of the proposed Pro-tuning could be attributed to its capability of modeling the task-specific prior for the latent manifold of image data. Essentially, the pre-trained vision model learns a data manifold. When various perturbations occur on a downstream dataset, the downstream data will deviate from the learned data manifold. To alleviate this problem, traditional transfer methods typically finetune the whole learned data manifold on the downstream dataset. However, due to a limited amount of downstream data, this way is prone to 1) overfitting on the downstream dataset; 2) extreme difficulty in optimization, as the learned data manifold could be a fine-detailed structure with high-level semantics. By learning a task-specific prior, our Pro-tuning can guide the deviated downstream data back to the learned manifold, thus avoiding the performance drop of the pre-trained model on downstream tasks.

	\section{Conclusion}
	\label{conclusion}
	In this work, we present Pro-tuning, a parameter-efficient vision tuning paradigm by extending the prompt-based philosophy to pre-trained vision models. The main idea of Pro-tuning is to maximally mitigate the gap between pre-training and tuning on downstream tasks by adapting the feature distribution of the pre-trained model at various levels. To this end, we propose multiple stage-wise prompt blocks, which can be simply plugged into given frozen pre-trained models, \textit{e.g.}, CNNs and vision transformers. Extensive experiments on fifteen challenging vision datasets have demonstrated the effectiveness and generalizability of our method.
	
	\textit{Limitation.}
	The limitations of Pro-tuning are mainly three aspects:~1) it is applicable in settings where the categories lie in a closed set; 2) it relies on a strong pre-trained model; 3) the inference still has the full computational cost of pre-trained models, which may limit its development in the edge devices.
	
	\textit{Broader impact.}
	Our work paves the new road to applying the pre-trained vision models. It has a broad application prospect in the common scenarios, such as cloud services. However, the proposed method may be maliciously used to solve those downstream tasks of involving damage peoples' livelihood or economic security, such as privacy theft and economic fraud. This issue warrants further consideration when this work is deployed in publicly available platforms such as cloud services.
	
	\bibliographystyle{plain}
	\bibliography{NeurIPS2022.bib}

\end{document}